\documentclass{article}

     \usepackage[nonatbib, preprint]{neurips_2024}

\usepackage[utf8]{inputenc} 
\usepackage[T1]{fontenc}    
\usepackage{hyperref}       
\usepackage{url}            
\usepackage{booktabs}       
\usepackage{amsfonts}       
\usepackage{nicefrac}       
\usepackage{microtype}      
\usepackage{xcolor}         
\usepackage[english]{babel}
\usepackage[numbers]{natbib}

\usepackage{multirow}
\usepackage{amsmath}
\usepackage{graphicx}
\usepackage{subfig}
\usepackage{hyperref}

\title{Talk to Parallel LiDARs: A Human-LiDAR Interaction Method Based on 3D Visual Grounding}

\author{%
  Yuhang Liu \\
  Institute of Automation\\
  Chinese Academy of Sciences\\
  Beijing, 100049 \\
  \texttt{liuyuhang2021@ia.ac.cn} \\
  \And
  Boyi Sun \\
  Institute of Automation \\
  Chinese Academy of Sciences \\
  Beijing, 100049 \\
  \texttt{sunboyi2024@ia.ac.cn} \\
  \AND
  Guixu Zheng \\
  College of Electronic Engineering \\
  South China Agricultural University \\
  Guangzhou, 510000 \\
  \texttt{guixu@stu.scau.edu.cn} \\
  \And
  Yishuo Wang \\
  School of Computer Science \\
  Beijing Institute of Technology \\
  Beijing, 100081 \\
  \texttt{1320211115@bit.edu.cn} \\
  \And
  Jing Yang \\
  Institute of Automation\\
  Chinese Academy of Sciences\\
  Beijing, 100049 \\
  \texttt{yangjing2020@ia.ac.cn} \\
  \And
  Fei-Yue Wang \\
  Institute of Automation \\
  Chinese Academy of Sciences \\
  Beijing, 100049 \\
  \texttt{feiyue.wang@ia.ac.cn} \\
}

\begin{document}

\maketitle

\begin{abstract}
LiDAR sensors play a crucial role in various applications, especially in autonomous driving. Current research primarily focuses on optimizing perceptual models with point cloud data as input, while the exploration of deeper cognitive intelligence remains relatively limited. 
To address this challenge, parallel LiDARs have emerged as a novel theoretical framework for the next-generation intelligent LiDAR systems, which tightly integrate physical, digital, and social systems. To endow LiDAR systems with cognitive capabilities, we introduce the 3D visual grounding task into parallel LiDARs and present a novel human-computer interaction paradigm for LiDAR systems.
We propose Talk2LiDAR, a large-scale benchmark dataset tailored for 3D visual grounding in autonomous driving.
Additionally, we present a two-stage baseline approach and an efficient one-stage method named BEVGrounding, which significantly improves grounding accuracy by fusing coarse-grained sentence and fine-grained word embeddings with visual features. Our experiments on Talk2Car-3D and Talk2LiDAR datasets demonstrate the superior performance of BEVGrounding, laying a foundation for further research in this domain.
\end{abstract}

\section{Introduction}
\label{1}
Autonomous driving is experiencing rapid development, with high-performance sensing systems being a critical step towards achieving L4 or L5 autonomous driving \cite{10092278}. 
LiDAR sensors play a crucial role in vehicular sensing systems, which can collect point clouds with precise spatial information \citep{9455394}. 
However, current LiDAR systems suffer from the separation of software and hardware development, severely limiting the system's intelligence. 
LiDAR manufacturers prioritize hardware optimization, while autonomous driving companies focus solely on software development. 
To redefine and build a new generation of intelligent LiDAR systems, we propose the framework of parallel LiDARs as theoretical guidance and have constructed a prototype based on the DAWN experimental platform \citep{liu2022parallel, 10163892}. 
It tightly couples physical and virtual spaces, enabling joint optimization of sensing and perception links through virtual-real interaction. 
Previous work has explored leveraging perceptual information to enhance data utilization efficiency and found that it can also improve perceptual accuracy \citep{10163892}. 
However, a significant issue remains that these operating modes cannot cognize and reason about the scene, for example, anticipating potentially dangerous areas to avoid accidents. 
Therefore, this work introduces user instructions to grant cognitive abilities of the LiDAR system and focuses on the 3D visual grounding task in autonomous driving.

\begin{figure}[tb]
  \centering
  \includegraphics[width=\linewidth]{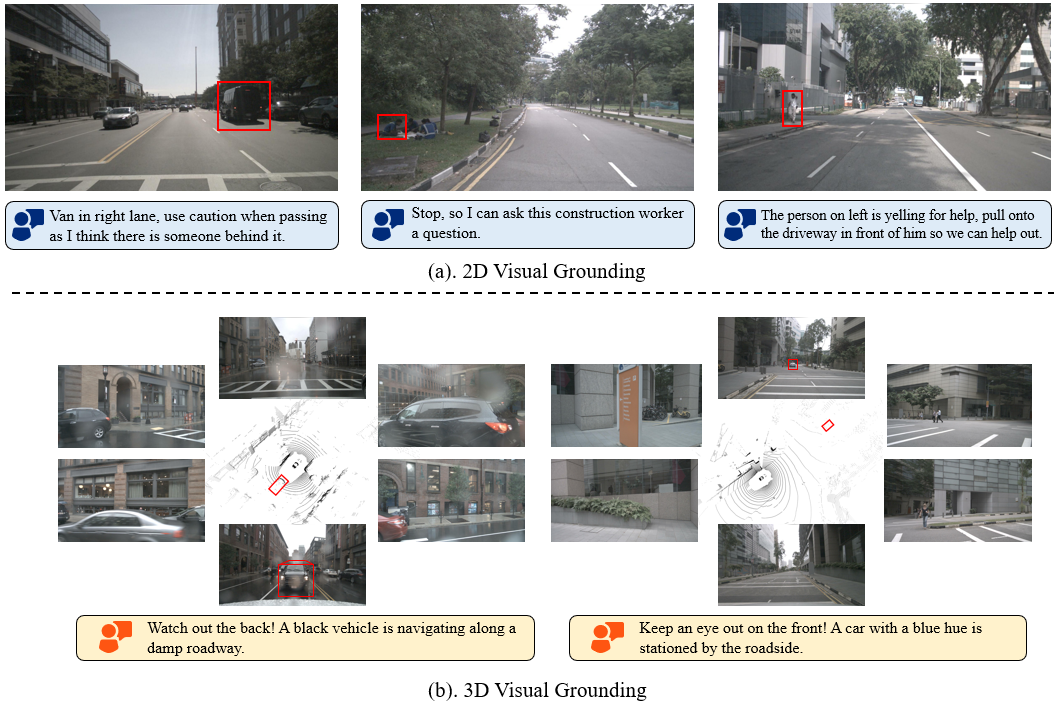}
  \caption{Introduction for the visual grounding task in autonomous driving. 2D visual grounding utilizes an image and language prompt as input (Fig.1a), while 3D visual grounding utilizes multi-view images, point clouds, and prompts as input (Fig.1b).}
  \label{intro}
\end{figure}

The 3D visual grounding task aims to identify the referred objects according to textual descriptions. It takes point clouds and text instructions as input and spits out 3D bounding boxes, which can be regarded as an image-based 2D visual grounding extension.  
Indoor 3D visual grounding has gained significant attention in recent years due to its promising applications in embodied intelligence \citep{lu2024scaneru}.
Several high-quality indoor datasets, such as ScanRefer \citep{chen2020scanrefer} and Sr3d \citep{achlioptas2020referit3d}, have been released to facilitate systematic research in this field. 
Following that, various methods have been developed to enhance grounding accuracy and they have already achieved notable progress \citep{luo20223d, jain2022bottom, wu2023eda}. 
However, existing works primarily focus on indoor environments with dense point clouds captured by RGBD sensors. 
They overlook outdoor scenarios with sparse LiDAR point clouds, which are crucial for applications like autonomous driving. 
Thus, there is an urgent need to investigate 3D visual grounding in the context of autonomous driving, as illustrated in Fig. \ref{intro}. This will pave the way for the development of an interactive guidance paradigm for parallel LiDARs.

This work advances the research of 3D visual grounding in driving scenarios, focusing on both dataset creation and method development.  
Concerning the dataset aspect, Talk2Car \citep{deruyttere2019talk2car} emerges as an outstanding dataset for 2D visual grounding in autonomous driving, enabling the association of matching point cloud data. However, its limited size, comprising 11,950 prompts across 7,818 frames, leads to poor generalizability.
To address this issue, we propose a novel large-scale 3D visual grounding dataset based on nuScenes \citep{caesar2020nuscenes}, named \textbf{\textit{Talk2LiDAR}}. 
It consists of 59,207 prompts across 6,419 scenes considering the diversity of viewpoint. 
To build our dataset cost-effectively, we introduce advanced MLLMs (Multimodal Large Language Models) \citep{NEURIPS2023_6dcf277e} and LLMs (Large Language Models) \citep{touvron2023llama} for automatic text prompt generation, followed by manual verification. 
In the method aspect, due to the scarcity of research in driving scenarios, we first propose a baseline approach. 
It adopts a two-stage processing pipeline based on the prevalent “detect-then-match” strategy. 
Then we introduce \textbf{\textit{BEVGrounding}}, a novel single-stage method that significantly boosts 3D visual grounding accuracy. 
Specifically, it utilizes a two-step fusion mechanism to achieve fine-grained alignment among text, image, and point cloud data.
Extensive experiments on Talk2Car-3D and Talk2LiDAR datasets effectively validate the superior performance of our proposed BEVGrounding method, laying a foundation for future research in this field.
The main contribution of this work can be summarized as follows:

\begin{itemize}
\item	
We innovatively introduce the 3D visual grounding task into parallel LiDARs, endowing LiDAR systems with cognitive capabilities through human-machine interaction. 
\item	
We propose \textbf{\textit{Talk2LiDAR}}, a new large-scale benchmark for 3D visual grounding in autonomous driving. 
\item	
We develop a two-stage baseline approach and an efficient one-stage method, called \textbf{\textit{BEVGrounding}}. It utilizes coarse-grained sentence and fine-grained word embeddings to fuse visual and textual features.  
\end{itemize}


\section{Related Work}
\subsection{Parallel LiDARs}
Parallel LiDAR emerges as a novel class of intelligent 3D sensors built upon parallel intelligence which is capable of capturing both physical and semantic information of 3D scenes \citep{liu2022parallel, liu2022radarverses}.
Parallel intelligence is an innovative theoretical framework proposed by Prof. Wang in 2004 \citep{wang2004parallel}.  
It integrates cyber, physical, and social spaces for intelligent systems catering to biological humans, robots, and digital humans \citep{5552591}. 
Currently, it has garnered widespread attention and found applications in various fields, such as autonomous driving \citep{8039015, 9956903}, sensing \citep{9970442, 10044575}, and manufacturing \citep{9970436}.
To facilitate research on parallel sensing, we have established a comprehensive experimental platform, \textbf{DAWN}, short for \textbf{D}igital \textbf{A}rtificial \textbf{W}orld for \textbf{N}atural. It supports exploration in various sub-projects such as parallel LiDARs and parallel light fields. 
Parallel LiDAR was originally proposed in \citep{liu2022parallel} which consists of three main parts: descriptive, predictive, and prescriptive LiDARs. 
Descriptive LiDARs focus on constructing digital LiDAR representations; predictive LiDARs emphasize the importance of computational experiments in cyberspace; while prescriptive LiDARs facilitate real-time interaction between the physical and digital LiDAR systems. 
\citep{10163892} proposed a software-defined parallel LiDAR model and constructed a hardware prototype in the DAWN platform.
It allows for dynamic adjustment of laser beam distribution to optimize the utilization of sensing resources.
To provide more comprehensive information, \citep{liu2024hpl} presented a novel HPL-ViT method to enhance the perception accuracy of heterogeneous parallel LiDARs in V2V.
Furthermore, \citep{9874864} discusses accurate modeling of parallel LiDARs in adverse weather. 
This paper delves into the 3D visual grounding task, aiming to further refine sensing resource allocation through human-LiDAR interaction.

\subsection{Visual Grounding in Autonomous Driving}
Visual grounding plays a critical role in autonomous driving by facilitating efficient human-computer interaction between drivers and vehicles \citep{deruyttere2020commands}. 
Current research primarily concentrates on 2D object detection and tracking based on language references using images or videos. 
The Talk2Car dataset, built upon nuScenes, serves as a pioneering benchmark that introduces the visual grounding task within autonomous driving \citep{deruyttere2019talk2car}. 
Numerous advancements have been made in enhancing visual grounding accuracy, with significant progress achieved \citep{dai2020commands, luo2020c4av, rufus2020cosine}. 
\citep{grujicic2022predicting} has extended the Talk2Car dataset to Talk2Car-Destination, enabling language-guided destination prediction. 
\citep{zhan2024mono3dvg} introduced a novel 3D visual grounding task utilizing a single image as input and established the Mono3DRefer dataset for evaluation. 
However, these approaches are limited to grounding individual objects, which falls short of capturing the complexities of real-world environments. 
To address this limitation, \citep{wu2023referring} proposed the ReferKITTI dataset, enabling grounding multiple objects with a single prompt.
Building upon this work, \citep{wu2023language} constructed the Nuprompt dataset and introduced PromptTrack, a method that leverages multi-view images for 3D tracking of referred objects. 
Notably, MSSG \citep{cheng2023language} stands as the sole approach that incorporates LiDAR point clouds for 3D visual grounding. 
Nevertheless, it suffers from limited details on the experimental setup and evaluation metrics, hindering replication by subsequent researchers.
This work delves into both data and methodological aspects of the 3D visual grounding task in autonomous driving, establishing a solid foundation for future research.

\subsection{3D Visual Grounding}
3D visual grounding aims to pinpoint objects according to the user's textual descriptions. 
Compared to complex outdoor environments, indoor settings have received more research attention due to their simpler scene structures.
Multiple datasets like ScanRefer \citep{chen2020scanrefer}, Sr3d, and Nr3d \citep{achlioptas2020referit3d} have been released to provide robust evaluation benchmarks for indoor 3D visual grounding.
Two-stage methods have dominated this landscape, achieving promising results in indoor scenes. 
These methods typically utilize a pre-trained object detector to generate candidate regions and extract prompt embeddings through a frozen text encoder.
The second stage focuses on matching the proposals with textual features to identify the final referred object.
\citep{he2021transrefer3d, zhao20213dvg} employ self-attention and cross-attention mechanisms for improved feature fusion, and \citep{yang2021sat, cai20223djcg} incorporate additional image information. 
However, a crucial limitation lies in their inability to recover missed objects during the initial stage.
To address this issue, single-stage methods have emerged and demonstrated competitive results on public datasets.
3D-SPS \citep{luo20223d} stands as the pioneer single-stage method, utilizing text features to progressively guide key point selection. 
Similarly, BUTD-DETR \citep{jain2022bottom} leverages a Transformer decoder for referred object prediction.
Recent advancements like EDA \citep{wu2023eda} introduce a text decoupling module, enabling finer-grained alignment by decomposing sentences into semantic components.
While these methods hold significant promise for indoor environments, they often overlook the vast potential of outdoor applications, particularly in autonomous driving.
This work bridges this gap by proposing BEVGrounding, a novel method specifically designed for 3D visual grounding in autonomous driving.

\section{Talk2LiDAR Dataset}
\label{3}

In this section, we will first introduce the statistics of our proposed Talk2LiDAR dataset. Then we present its construction process, highlighting the role of foundation models in its development.

\subsection{Dataset Statistics}
\label{3.1}

The Talk2LiDAR dataset is the first large-scale dataset specifically designed for LiDAR-based 3D visual grounding in autonomous driving. It’s built on the nuScenes dataset, a classic autonomous driving dataset collected in Boston and Singapore. 
 Talk2LiDAR establishes itself as the largest benchmark for 3D visual grounding in autonomous driving. It features 59,207 language prompts across 6,419 scenes, with an average of 9.2 prompts per scene. 
Notably, Talk2LiDAR surpasses prior datasets confined to front-view images.
Referred objects are distributed around the ego vehicle, viewable from six distinct image perspectives. It offers a more comprehensive and realistic representation of real-world driving scenarios.
Assisted by advanced foundation models \citep{NEURIPS2023_6dcf277e, touvron2023llama}, Talk2LiDAR boasts a diverse vocabulary within its language prompts, revealing its rich incorporation of information regarding location, category, and color.
Additional details are provided in App.\ref{a11}.

\subsection{Dataset Construction}
\label{3.2}

Prior visual grounding datasets relied heavily on the manual generation of language prompts, leading to significant time consumption and labor costs. Additionally, manual annotation often led to an abundance of repetitive words, hindering vocabulary diversity. 
To address these limitations, we introduce a novel three-step data annotation pipeline assisted by powerful foundation models. 
By leveraging the cognitive capabilities of these models, we significantly reduce the workload associated with creating triplet data pairs (text, image, and point cloud). A discussion of each step is provided below, with more details available in App.\ref{a12}. 

{\bf{Step 1:}}
Although Talk2LiDAR focuses on LiDAR-based 3D visual grounding, we describe referred objects using multi-view images, mirroring how drivers perceive their environment. 
Since the nuScenes dataset contains multiple continuous segments, we randomly sample 20\% of all frames to eliminate redundancy. 
Subsequently, we visualize the annotations on the images and retain only those with complete bounding boxes. 
Finally, to ensure high dataset quality, we manually filter out any remaining samples with ambiguities or noticeable discrepancies.

{\bf{Step 2:}}
Following manual filtering, we utilize the advanced multimodal foundation model LLaVA \citep{NEURIPS2023_6dcf277e} to automatically generate initial textual descriptions.
We feed a well-designed prompt along with the images into LLaVA to generate descriptions of the referred objects. 
However, we observe that LLaVA tends to prioritize prominent objects in the image or macroscopic factors like weather conditions.
To address this issue, we refine the prompts to steer LLaVA towards a stronger focus on the objects in the bounding boxes.

{\bf{Step 3:}}
We have obtained preliminary image-point cloud-text triplet data pairs by step 2. 
However, foundation models often employ probabilistic token prediction, leading to repetitive phrasings and a limited vocabulary in the generated prompts.
To address this limitation, we utilize the latest language foundation model, LLaMA2 \citep{touvron2023llama}, for refinement, aiming to enhance the diversity of descriptions.
Additionally, we integrate viewpoint information into the descriptions to incorporate more comprehensive spatial information.

\section{Methods}
\label{4}

\begin{figure}[!t]
\centering
\includegraphics[width=5in]{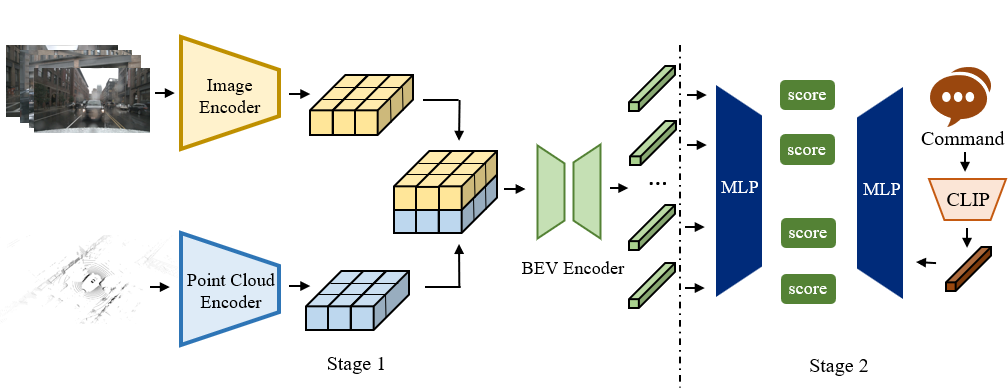}
\caption{The architecture of the two-stage baseline method.}
\label{baseline}
\end{figure}

\subsection{Problem Definition}
LiDAR-based 3D visual grounding presents a novel and promising task in autonomous driving. Given a textual instruction $T$, point cloud data $P$, and images $I$, the objective is to precisely localize the referred object $B$ within the 3D scene. 
$T$ is a textual description consisting of $L$ words, formally represented as $T=\{w_1, ..., w_L\}$.
$P = \{p_1, …, p_n\}$ is a single frame of point cloud data.
Each point $p_i (i=1, ..., n)$ is a 4-tuple $(x_i, y_i, z_i, i_i)$, where $x_i$, $y_i$, $z_i$ specifies its spatial coordinates and $i_i$ represents its intensity information. 
Image set $I=\{I_1, …, I_6\}$ comprises a collection of six images captured from different viewpoints surrounding the vehicle.
The 3D bounding box $B$ of the referred object is represented as $(x, y, z, l, w, h, \alpha)$. Here, $x$, $y$, $z$ denotes the center location of the bounding box, $l$, $w$, $h$ represents its dimensions, and $\alpha$ corresponds to its rotation angle.

\subsection{Baseline}
\label{4.2}
Given the nascent state of 3D visual grounding research in autonomous driving, we establish a baseline method adhering to the dominant two-stage paradigm.
As illustrated in Fig. \ref{baseline}, the baseline method utilizes BEVFusion \citep{10160968} for 3D object detection in the first stage, generating candidate proposals along with their extracted features. 
The second stage leverages a pre-trained language encoder to extract sentence-level embeddings from the textual description.
We then introduce a lightweight matching network to identify the referred object.
Specifically, the object and language features are fed into separate MLPs for feature alignment in the matching network.
We then compute the final score using matrix multiplication and select the candidate object with the highest score as the text-referred one:

\begin{equation}
s = E_{l}(f_{lang}) * E_{o}(f_{obj})
\end{equation}

$E_{l}$ and $E_{o}$ represent the MLPs for language and object features, each consisting of two fully connected layers. 
$*$ denotes the matrix multiplication operation, and $s$ represents the final score for each candidate object. 
During training, the weights of the language encoder are frozen, while only the weights of the lightweight matching network are updated. 
The cross-entropy loss function is employed for optimization.

\subsection{BEVGrounding}

\begin{figure*}[!t]
\centering
\includegraphics[width=5in]{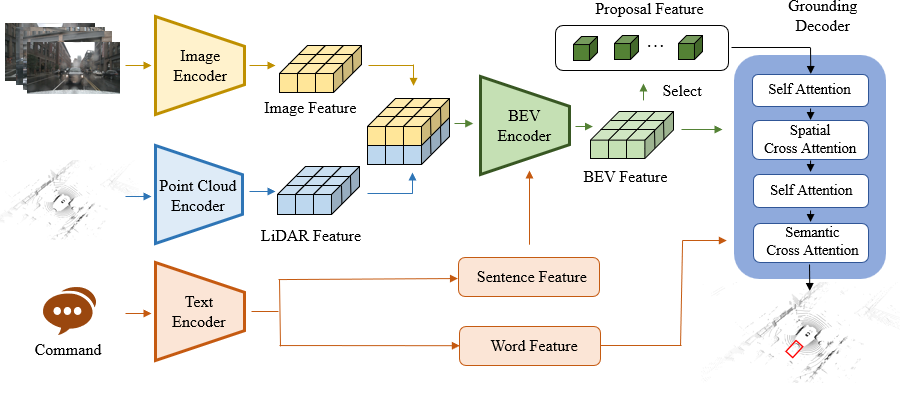}
\caption{The network architecture of our proposed one-stage BEVGrounding method.}
\label{bevgrounding}
\end{figure*}

The baseline method adopts a common two-stage approach, while it poses challenges for practical applications due to its training and deployment complexities. 
To address this issue, we propose BEVGrounding, a novel one-stage method designed for 3D visual grounding in autonomous driving. The overall architecture of BEVGrounding is shown in Fig. \ref{bevgrounding}. 

\subsubsection{Unimodal Encoder}
BEVGrounding, being a multimodal method, leverages images, point clouds, and text data as input. 
The point cloud branch employs a grid-based encoder $E_P$ to address the high computational cost associated with point-based encoders. 
We utilize sparse convolution to extract voxel features, which are subsequently flattened to obtain $f_{bev}^P$.
The image branch utilizes a Swin-Transformer architecture as the encoder $E_I$ to extract features from multi-view images.
These features are then projected onto the BEV (Bird's-Eye View) space, resulting in $f_{bev}^I$.
For the text instruction, we experiment with CLIP \cite{radford2021learning} as the text encoder $E_T$ to learn both word-level and sentence-level embeddings, facilitating a more fine-grained feature interaction between different modalities:

\begin{equation}
f_{bev}^P = E_P(P); f_{bev}^I = E_I(I); f_{sen}, f_{word} = E_T(T)
\end{equation}

\subsubsection{Trimodal BEV Encoder}

Following the extraction of individual modality features, we design a trimodal BEV encoder for global feature fusion. 
Considering the real-time requirements, our BEV encoder leverages a purely CNN architecture, eschewing the use of a Transformer-based approach. 
We first perform a preliminary fusion of $f_{bev}^P$ and $f_{bev}^I$ to obtain the initial fused representation, denoted as $f_{bev}^{'}$. 
Subsequently, $f_{sen}$ obtained from the text encoder is broadcasted to match the spatial dimensions of the BEV feature maps and then concatenated with $f_{bev}^{'}$.
To alleviate the computational burden, a 1x1 convolution layer is employed to reduce the number of feature channels. Finally, we adopt a classic feature pyramid network to achieve a global coarse-grained fusion, resulting in the representation $f_{bev}^{''}$.

\subsubsection{Grounding Decoder}

Guided by the heatmap scores, we select a subset of proposals from $f_{bev}^{''}$ that exhibit a high likelihood of corresponding to the text-referred objects.
These selected proposal features are then fed into the grounding decoder for further fine-grained feature fusion. 
Our proposed grounding decoder consists of four key blocks: two self-attention (SA) blocks, a spatial cross-attention (SPCA) block, and a semantic cross-attention (SECA) block. 
Each block consists of a multi-head attention layer and a FFN layer. 
The SA block effectively captures the global dependencies within the proposal features. 
The SPCA block fuses proposal features $f_{pro}$ and $f_{bev}^{''}$, while SECA facilitates interaction between $f_{pro}$ and the word-level embeddings $f_{word}$, providing more fine-grained textual features for candidate objects. The formulation of the attention layer in each module is as follows: 

\begin{equation}
SA(f_{pro}) = \sigma(Q(f_{pro})K(f_{pro}^T))V(f_{pro})
\end{equation}
\begin{equation}
SPCA(f_{pro}, f_{bev}^{''}) = \sigma(Q(f_{pro})K(f_{bev}^{''}))V(f_{bev}^{''})
\end{equation}
\begin{equation}
SECA(f_{pro}, f_{word}) = \sigma(Q(f_{pro})K(f_{word}))V(f_{word})
\end{equation}

Here, $\sigma$ is the softmax function. $Q$, $K$, and $V$ correspond to the query, key, and value transformation layer, respectively. 
Finally, we utilize a standard detection head to predict the 3D bounding boxes of the referring objects \citep{bai2022transfusion}. During the training phase, the loss function incorporates three key parts:

\begin{equation}
L_{all} = L_{heatmap} + L_{cls} + L_{reg}
\end{equation}

Consistent with prior work, $L_{heatmap}$ leverages Gaussian focal loss for proposal filtering, $L_{cls}$ uses the focal classification loss, and $L_{ref}$ adopts a L1 loss for bounding box regression. 
Furthermore, inspired by DETR \cite{carion2020end}, we incorporate the Hungarian algorithm for bipartite matching during the training process

\section{Experiments}

\subsection{Implementation Details}
\label{5.1}

We conduct experiments on both the Talk2Car-3D and our proposed Talk2LiDAR datasets. Talk2Car-3D is derived from the original Talk2Car through a three-step preprocessing procedure. 
First, we categorize the referred objects into 10 standard categories according to common 3D object detection conventions. 
Second, we constrain their positions, retaining only objects in the range requirement of [-54, -54, -5] \textless [x, y, z] \textless [54, 54, 3], where [x, y, z] represent the center location. 
Third, we filter objects based on the number of point clouds inside each object, keeping only those with at least one point. 
The processed Talk2Car-3D comprises 8,352 data frames. Following the original Talk2Car split, we utilize 7,332 frames for training and the remaining 1,020 frames for testing. 
To provide additional context for referred objects, we define two attribute labels for each sample: \textit{"unique"} and \textit{"multiple"}.
 The \textit{“unique”} label indicates that the referred object is the only category-matched target in the frame, while \textit{“multiple”} signifies the presence of several objects in the same category. 
The training and testing sets of Talk2Car-3D contain 836 and 106 frames with the \textit{“unique”} attribute, respectively. 
We apply similar processing steps to the Talk2LiDAR dataset. The processed Talk2LiDAR training set comprises 48,813 frames, while the testing set comprises 12,394 frames. Within its training and testing sets, 2,344 and 694 frames are labeled as \textit{"unique"}, respectively.
Additional details are provided in App.\ref{a21}.

\subsection{Quantitative Analysis}
\label{5.2}

We design multiple two-stage methods to enable more fair evaluations and compare them with our proposed BEVGrounding.
\textbf{GT-Rand} randomly selects a ground truth box as the prediction result, while \textbf{Pred-Rand} randomly selects a predicted proposal as the referred object. 
\textbf{Pred-Best} chooses the candidate with the highest confidence score among all detection boxes. \textbf{Baseline} is the method we proposed in Sec. \ref{4.2} and \textbf{-L} denotes that only point cloud data is used. 

\subsubsection{Performance on Talk2Car-3D}

\begin{table*}[htbp]
\centering
\caption{Comparasion with other methods on Talk2Car-3D.}
\scalebox{0.66}{
\begin{tabular}{cccccccccc}
\hline

\multirow{2}{*}{}    & \multirow{2}{*}{Method} & \multirow{2}{*}{Type} & \multicolumn{2}{c}{Unique} & \multicolumn{2}{c}{Multiple} & \multicolumn{2}{c}{Overall} \\
                     &                       &  &  Acc@0.25 (\%)      & Acc@0.5 (\%)      & Acc@0.25 (\%)       & Acc@0.5 (\%)       & Acc@0.25 (\%)      & Acc@0.5 (\%)       \\ 
\hline

\multirow{7}{*}{BEV} & GT-Rand  & Two-Stage           & 9.43      & 9.43      & 6.78       & 6.78      & 7.06      & 7.06      \\
                     & Pred-Rand            & Two-Stage      & 6.60   & 4.72       & 7.44  & 5.25           & 7.35  & 5.20     \\
& Pred-Best               & Two-Stage             & 0.94   & 0.94         & 14.33  & 12.91           & 12.94  & 11.67           \\
& Baseline-L              & Two-Stage             & 2.83   & 1.89          & 27.46  & 26.59           & 24.90  & 24.02           \\
& Baseline                & Two-Stage             & 13.21  & 9.43          & 30.53  & 27.35           & 28.73  & 25.49           \\
& \textbf{BEVGrounding-L} &   \textbf{One-Stage}   &   \textbf{32.08 (+18.87)}  & \textbf{16.98 (+7.55)}         & \textbf{43.76 (+13.23)}  & \textbf{28.12 (+0.77)}           & \textbf{42.55 (+13.82)}  & \textbf{26.96 (+1.47)}           \\
& \textbf{BEVGrounding}            & \textbf{One-Stage}             & \textbf{33.02 (+19.81)}  & \textbf{17.92 (+8.49)}         & \textbf{45.30 (+14.77)}  & \textbf{29.76 (+2.41)}           & \textbf{44.02 (+15.29)}  & \textbf{28.53 (+3.04)}           \\

\hline

\multirow{7}{*}{3D} & GT-Rand                 & Two-Stage            & 9.43   & 9.43         & 6.78   & 6.78            & 7.06   & 7.06            \\
& Pred-Rand               & Two-Stage             & 6.60   & 1.89          & 7.11  & 3.94            & 7.06   & 3.73        \\
& Pred-Best               & Two-Stage             & 0.94   & 0.94          & 14.11  & 12.69           & 12.75  & 11.47          \\
& Baseline-L              & Two-Stage             & 2.83   & 0.94          & 25.38  & 21.77          & 23.04  & 19.61           \\
& Baseline                & Two-Stage             & 13.26  & 7.55         & 28.01  & 26.48           & 26.37  & 24.51          \\
& \textbf{BEVGrounding-L} & \textbf{One-Stage}             & \textbf{24.53 (+11.27)}  & \textbf{7.55 (-)}         & \textbf{36.32 (+8.31)}  & \textbf{18.38 (-)}           & \textbf{35.10 (+8.73)}  & \textbf{17.25 (-)}          \\
& \textbf{BEVGrounding}            & \textbf{One-Stage}             & \textbf{25.57 (+12.31)}  & \textbf{8.49 (-)}          & \textbf{37.64 (+9.63)}  & \textbf{20.90 (-)}           & \textbf{36.37 (+10.00)}  & \textbf{19.61 (-)}          \\

\hline
\end{tabular}}
\label{talk2car}
\end{table*}

Tab.\ref{talk2car} presents the accuracy of all methods from BEV and 3D perspectives. The overall performance of GT-Rand and Pred-Rand is poor, with accuracies below 10\%, effectively demonstrating the difficulty of the 3D visual grounding task. 
Although Pred-Best shows slight improvement, it still exhibits significant randomness due to the lack of textual features. 
The baseline significantly enhances the accuracy metrics compared to the aforementioned methods.
BEVGrounding outperforms all other methods on almost all metrics, except for the 3D Acc@0.5. 
Notably, it achieves 44.02\% on BEV Acc@0.25, exceeding the second-best method by 15.29\%.
BEVGrounding-L, which utilizes only point cloud data as input, significantly outperforms the multimodal fusion baseline, demonstrating the superiority of our single-stage architecture.
However, it experiences a slight decrease in 3D Acc@0.5.
We speculate that it could be attributed to BEVGrounding's emphasis on semantic alignment without fully considering fine-grained geometric features, leading to negative effects at higher IoU thresholds.
Furthermore, we observe an anomaly where samples labeled as \textit{“unique”}, intuitively simpler, exhibit lower accuracy. 
Additional analyses and details are provided in App.\ref{a22}.

\subsubsection{Performance on Talk2LiDAR}

\begin{table*}[htbp]
\centering
\caption{Comparasion with other methods on Talk2LiDAR.}
\scalebox{0.66}{\begin{tabular}{cccccccccc}
\hline

\multirow{2}{*}{}    & \multirow{2}{*}{Method} & \multirow{2}{*}{Type} & \multicolumn{2}{c}{Unique} & \multicolumn{2}{c}{Multiple} & \multicolumn{2}{c}{Overall} \\
                     &                       &  &  Acc@0.25 (\%)      & Acc@0.5 (\%)      & Acc@0.25 (\%)       & Acc@0.5 (\%)       & Acc@0.25 (\%)      & Acc@0.5 (\%)       \\ 
\hline

\multirow{7}{*}{BEV} & GT-Rand  & Two-Stage           & 6.20      & 6.20      & 3.79       & 3.79      & 3.88      & 3.88      \\
                     & Pred-Rand            & Two-Stage      & 5.48   & 4.47       & 4.44  & 3.85           & 4.49  & 3.89     \\
& Pred-Best               & Two-Stage             & 2.88   & 2.59         & 6.95  & 6.79           & 6.72  & 6.55           \\
& Baseline-L              & Two-Stage             & 4.76   & 4.32          & 8.42  & 8.17           & 8.21  & 7.96           \\
& Baseline                & Two-Stage             & 9.22  & 8.07          & 10.68  & 10.32           & 10.61  & 10.20           \\
& \textbf{BEVGrounding-L} &   \textbf{One-Stage}   &   \textbf{14.12 (+4.90)}  & \textbf{12.97 (+4.90)}         & \textbf{16.39 (+5.71)}  & \textbf{11.97 (+1.65)}           & \textbf{16.27 (+5.66)}  & \textbf{12.02 (+1.82)}           \\
& \textbf{BEVGrounding}            & \textbf{One-Stage}             & \textbf{15.99 (+6.77)}  & \textbf{14.99 (+6.92)}         & \textbf{18.19 (+7.51)}  & \textbf{13.39 (+3.07)}           & \textbf{18.07 (+7.46)}  & \textbf{13.48 (+3.28)}           \\

\hline

\multirow{7}{*}{3D} & GT-Rand                 & Two-Stage            & 6.20      & 6.20      & 3.79       & 3.79      & 3.88      & 3.88            \\
& Pred-Rand               & Two-Stage             & 5.48   & 3.31          & 4.28  & 3.35            & 4.35   & 3.35        \\
& Pred-Best               & Two-Stage             & 2.88   & 2.31          & 6.91  & 6.48           & 6.68  & 6.24          \\
& Baseline-L              & Two-Stage             & 4.76   & 3.46          & 8.34  & 7.94          & 8.62  & 7.69           \\
& Baseline                & Two-Stage             & 8.65  & 7.35         & 10.05  & 9.67           & 9.97  & 9.54          \\
& \textbf{BEVGrounding-L} & \textbf{One-Stage}             & \textbf{13.83 (+5.18)}  & \textbf{6.77 (-)}         & \textbf{16.13 (+6.08)}  & \textbf{7.68 (-)}           & \textbf{16.00 (+6.03)}  & \textbf{7.63 (-)}          \\
& \textbf{BEVGrounding}            & \textbf{One-Stage}             & \textbf{15.27 (+6.62)}  & \textbf{8.65 (+1.30)}          & \textbf{17.49 (+7.44)}  & \textbf{8.72 (-)}           & \textbf{17.36 (+7.39)}  & \textbf{8.71 (-)}          \\

\hline
\end{tabular}}
\label{talk2lidar}
\end{table*}

Tab.\ref{talk2lidar} illustrates the results of all methods on the Talk2LiDAR dataset. 
Compared to their performance on Talk2Car-3D, all methods show a significant drop in accuracy. 
This decline can be attributed to the increased complexity of language prompts and object locations in Talk2LiDAR, indicating that there is substantial room for improvement in future research.
Among all the methods, BEVGrounding stands out by achieving SOTA performance on most evaluation metrics, with an average improvement of 5\%-7\%.
However, it also demonstrates a slight decrease in 3D Acc@0.5, consistent with the trend observed on Talk2Car-3D.

\subsubsection{Ablation Studies}

\begin{figure*}[!t]
\centering
\includegraphics[width=5.5in]{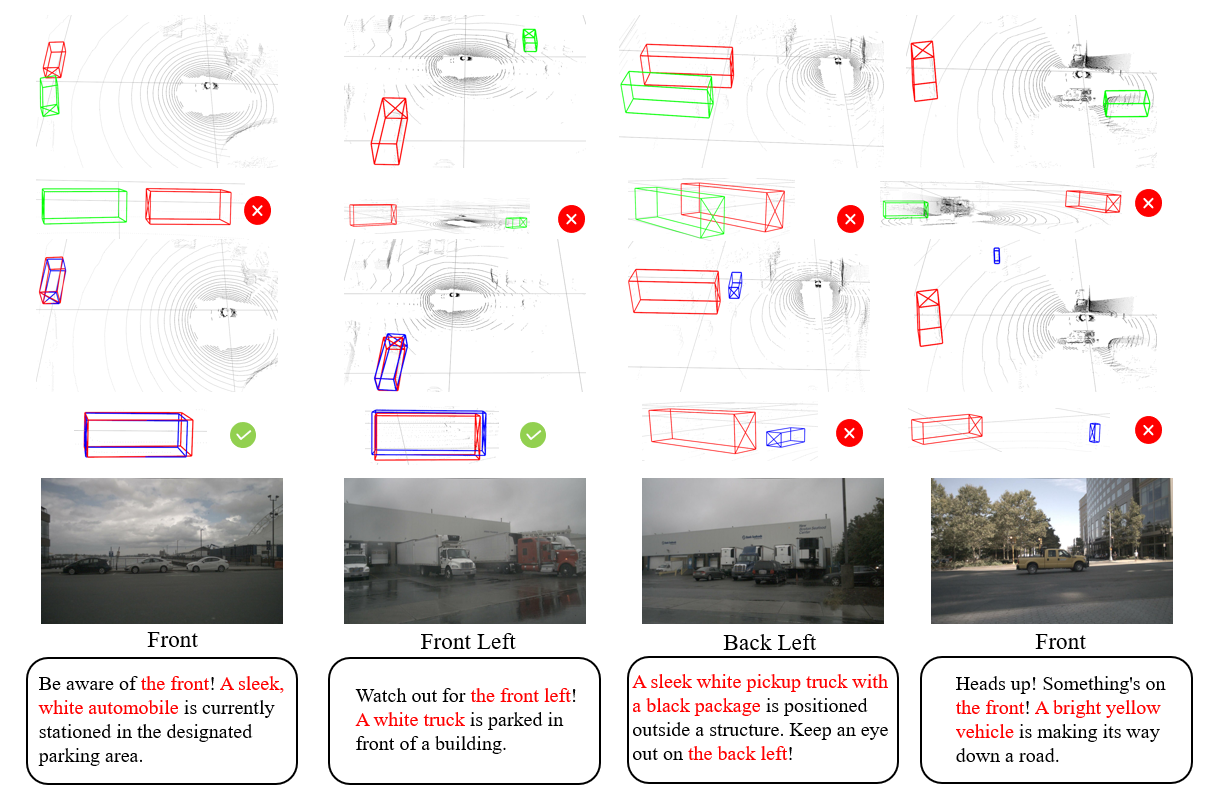}
\caption{Visualization results of the two-stage baseline and one-stage BEVGrounding method. {\color{red}Red}, {\color{blue}blue}, and {\color{green}green} boxes denote the ground truth, predicted boxes by the baseline, and predicted boxes by BEVGrounding, respectively.}
\label{qa}
\end{figure*}

\begin{table*}[htbp]
\centering
\caption{Ablation studies on text encoder.}
\scalebox{0.66}{\begin{tabular}{cccccccc}
\hline
\multirow{2}{*}{}    & \multirow{2}{*}{Text Encoder} & \multicolumn{2}{c}{Unique} & \multicolumn{2}{c}{Multiple} & \multicolumn{2}{c}{Overall} \\
                     &                       & Acc@0.25 (\%)      & Acc@0.5 (\%)      & Acc@0.25 (\%)       & Acc@0.5 (\%)       & Acc@0.25 (\%)      & Acc@0.5 (\%)       \\ \hline
\multirow{4}{*}{BEV} & GloVe-50b             & 16.98      & 8.49      & 28.99       & 22.65      & 27.75      & 21.18      \\
                     & GloVe-100b            & 27.36      & 12.26     & 32.82       & 23.96      & 32.25      & 22.75     \\
                     & GloVe-200b            & 25.47      & 8.49      & 37.20       & 25.93      & 35.98      & 24.12      \\
                     & \textbf{CLIP}                  & \textbf{33.02}      & \textbf{17.92}     & \textbf{45.30}       & \textbf{29.76}      & \textbf{44.02}      & \textbf{28.53}       \\
\hline
\multirow{4}{*}{3D}  & GloVe-50b             & 16.04     & 2.83      & 27.24       & 15.75      & 26.08      & 14.41      \\
                     & GloVe-100b            & 16.98      & 3.77      & 28.56       & 16.63      & 27.32      & 15.29      \\
                     & GloVe-200b            & 16.04      & 3.77      & 33.48       & 18.71      & 31.67      & 17.16      \\
                     & \textbf{CLIP}                  & \textbf{25.41}      & \textbf{8.49}      & \textbf{37.64}       & \textbf{20.90}      & \textbf{36.37}      & \textbf{19.61}      \\ \hline
\end{tabular}}
\label{textencoder}
\end{table*}

\begin{table}[htbp]
\centering
\caption{Ablation studies on BEVGrounding's module.}
\scalebox{0.66}{\begin{tabular}{ccccccc}
\hline
\multirow{2}{*}{EN} & \multirow{2}{*}{SPCA} & \multirow{2}{*}{SECA} & \multicolumn{2}{c}{Overall@BEV} & \multicolumn{2}{c}{Overall@3D} \\
                         &                                               &                                                 & Acc@0.25 & Acc@0.5 & Acc@0.25 & Acc@0.5 \\
\hline
                         &                                               &                                                 & 31.86 & 20.10 & 27.65 & 15.00 \\
\checkmark               &                                               &                                                 & 38.43 & 25.49 & 33.43 & 17.06 \\
\checkmark               & \checkmark                                    &                                                 & 37.84 & 24.90 & 31.37 & 16.96 \\
\checkmark               & \checkmark                                    & \checkmark                                      & \textbf{44.02} & \textbf{28.53} & \textbf{36.37} & \textbf{19.61} \\
\hline
\end{tabular}}
\label{module}
\end{table}

\begin{itemize}
\item{\textbf{Text Encoder: }We use CLIP as the language encoder in our experiments. However, prior research often utilizes a GRU-based language encoder with GloVE embeddings \cite{pennington2014glove}. 
To investigate the impact of different text encoders, we conduct ablation experiments on Talk2Car-3D and present the results in Tab.\ref{textencoder}. 
Our findings indicate that CLIP achieves superior performance, surpassing the GloVE-based approach by an average of 5\%. }
\item{\textbf{Module Components: }To assess the contribution of each module in BEVGrounding, we conduct ablation studies, and the results are detailed in Tab.\ref{module}.
It reveals that the encoder module exerts the most significant influence on the model's performance, leading to a 12.16\% accuracy improvement. 
Interestingly, SPCA and SECA modules exhibit comparable impacts, suggesting that both spatial and semantic feature interactions are crucial for crucial for accurate object grounding.
}
\end{itemize}

\subsection{Qualitative Analysis}
\label{5.3}

Fig. \ref{qa} illustrates the visualization results for both the baseline method and our proposed BEVGrounding approach. 
We can observe that one-stage BEVGrounding can generate more accurate 3D bounding boxes for the referred objects compared to the two-stage baseline. 
It stems from BEVGrounding's capability to effectively extract and integrate richer semantic information about the scene.
However, both methods still encounter numerous failures during testing, indicating significant room for improvement in 3D visual grounding for autonomous driving.

\section{Conclusions}
\label{6}
This work introduces the 3D visual grounding task into parallel LiDARs, aiming to equip sensors with a degree of cognitive capability.  
It provides a novel human-machine interaction approach for LiDAR systems. 
We establish the Talk2LiDAR dataset, a large-scale benchmark for 3D visual grounding, and propose BEVGrounding, a novel one-stage method that demonstrates promising results. 
Our future work will explore the integration of multimodal foundation models to further elevate the cognitive intelligence of parallel LiDAR systems.

\bibliography{ref}
\clearpage


\appendix

\section{Appendix}

\subsection{Talk2LiDAR Dataset}
\label{a1}

\subsubsection{Dataset Statistics}
\label{a11}

\begin{table*}[htbp]
  \centering
  \caption{A comprehensive comparison of visual grounding datasets.}
  \resizebox{\textwidth}{!}{%
  \begin{tabular}{*{11}{c}}
    \toprule
    Dataset & Publication & Scene & Scene Num. & Prompt Num. & Basic Tasks & Input Modality & Views\\
    \midrule
    ScanRefer \cite{chen2020scanrefer} &ECCV2020 &Indoor &800 &51583 &3D Det &PC &- \\
    Sr3d \cite{achlioptas2020referit3d} &ECCV2020 &Indoor &1273 &83572 &3D Det &PC &- \\
    Nr3d \cite{achlioptas2020referit3d} &ECCV2020 &Indoor &707 &41503 &3D Det &PC &- \\
    Multi3DRefer \cite{zhang2023multi3drefer} &ICCV2023 &Indoor &800 &61926 &3D Det &PC &- \\
    Talk2Car \cite{deruyttere2019talk2car} &EMNLP2019 &Outdoor &7818 &11959 &2D Det &Img &1 \\
    Refer-KITTI \cite{wu2023referring} & CVPR2023 &Outdoor &6650 &818 &2D MOT &Img &1 \\
    Mono3DRefer \cite{zhan2024mono3dvg} &AAAI2024 &Outdoor &2025 &41140 &3D Det&Img &1 \\
    NuPrompt \cite{wu2023language} &arXiv2023 &Outdoor &34149 &35367 &3D MOT &Img &6 \\
    \textbf{Talk2Car-3D} & \textbf{-} & \textbf{Outdoor} & \textbf{5534} & \textbf{8352} & \textbf{3D Det} & \textbf{PC, Img} & \textbf{1} \\
    \textbf{Talk2LiDAR} & \textbf{-} & \textbf{Outdoor} & \textbf{6419} & 
    \textbf{59207} & \textbf{3D Det} & \textbf{PC, Img} & \textbf{6} \\
    \bottomrule
  \end{tabular}%
  }
\label{vsdataset}
\end{table*}

\begin{figure}[!t]
\centering
\includegraphics[width=3in]{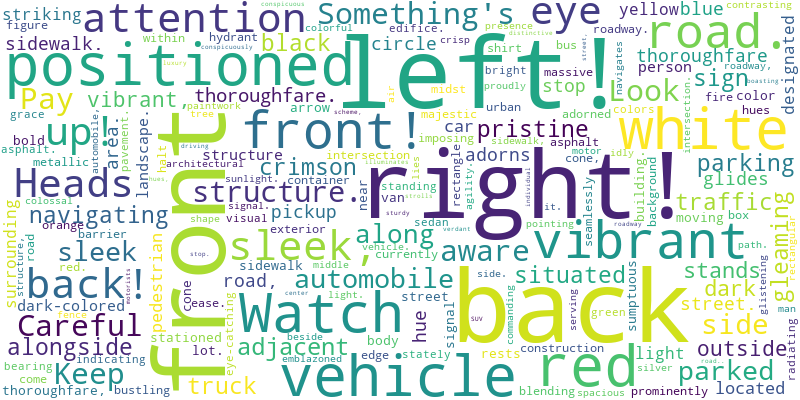}
\caption{Word cloud of language prompts from the Talk2LiDAR dataset.}
\label{word}
\end{figure}

Tab.\ref{vsdataset} presents a detailed comparison of Talk2LiDAR with other leading visual grounding datasets.
It's currently the largest dataset for visual grounding tasks in autonomous driving, featuring diversity in data modalities and viewpoints. 
Fig.\ref{word} shows a word cloud of language prompts from the Talk2LiDAR dataset. 
We can observe that it incorporates abundant information regarding location, category, and color characteristics.

\subsubsection{Dataset Construction}
\label{a12}

\begin{figure}[!t]
\centering
\includegraphics[width=5.5in]{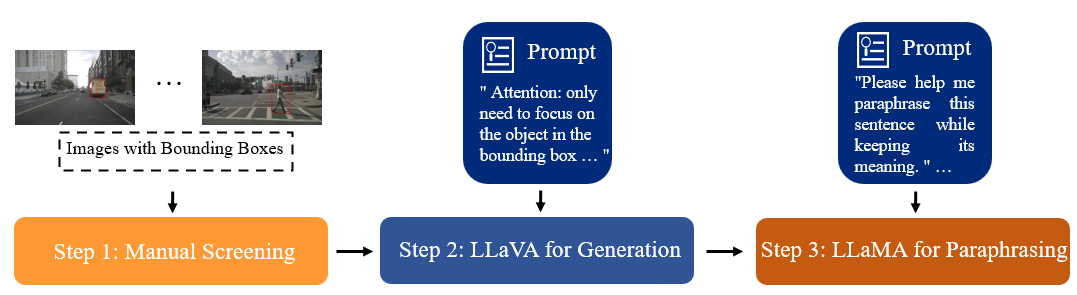}
\caption{The Talk2LiDAR dataset construction process.}
\label{construct}
\end{figure}

Fig.\ref{construct} illustrates the overall construction workflow of the Talk2LiDAR dataset. 

In step 2, we craft the following prompt to efficiently guide LLaVA to focus more on the referred objects in bounding boxes:

\begin{itemize}
\item{\textit{Attention: only need to focus on the object in the bounding box. Please use one or two sentences to describe the object in the red bounding box with greater detail, including its precise location, type, and color characteristics.}}
\end{itemize}

To enhance the diversity of descriptions, we provide LLaMA2 with the following prompts:

\begin{itemize}
\item{\textit{Please help me paraphrase this sentence while keeping its meaning.}}
\item{\textit{Please help me reword a sentence with richer vocabulary but keep its meaning.}}
\item{\textit{Help me reword a sentence, you should describe it in a different way.}}
\end{itemize}

We can obtain descriptions with a richer vocabulary after prompt paraphrasing. Below are some examples of the original (\textit{O}) and paraphrased (\textit{P}) prompts:

\begin{itemize}
\item{\textit{O: A car is driving down the street at night.}}
\item{\textit{R: Be aware of the back right! A luxuriant automobile is navigating the boulevard under the cover of darkness.}}
\item{\textit{O: A woman in a black dress is standing in the middle of a street.
}}
\item{\textit{R: Look out for the front! A statuesque woman in a sleek black gown stands majestically in the bustling street, her poise and elegance commanding attention.}}
\end{itemize}

\subsection{Experiments}
\label{a2}

\subsubsection{Implementation Details}
\label{a21}
We leverage the OpenPCDet framework for our experiments, with training performed on 8 Nvidia V100 GPUs.
For the baseline method, we first train the BEVFusion detection model and retain a maximum of 200 proposals. 
We then train the matching network for 20 epochs with a batch size of 4, using an SGD optimizer with a learning rate of 0.01.
For BEVGrounding, we begin by training the LiDAR branch for 20 epochs, after which we introduce the camera branch and fine-tune for an additional 6 epochs.
The Adam optimizer is employed, with learning rates of 0.001 and 0.0001 for these two stages, respectively.
Following prior work, we use accuracy with BEV or 3D IoU thresholds as our evaluation metrics, setting the thresholds at 0.25 and 0.5.

\subsubsection{Quantitative Analysis}
\label{a22}

\begin{figure}
  \centering
    \subfloat[]{
    		\includegraphics[scale=0.55]{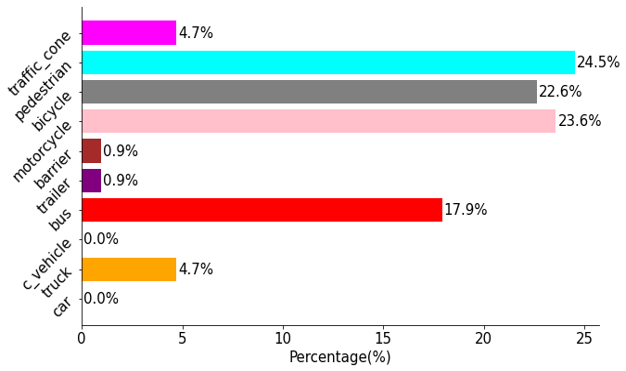}
            \label{fig:a}
            }
    \subfloat[]{
    		\includegraphics[scale=0.55]{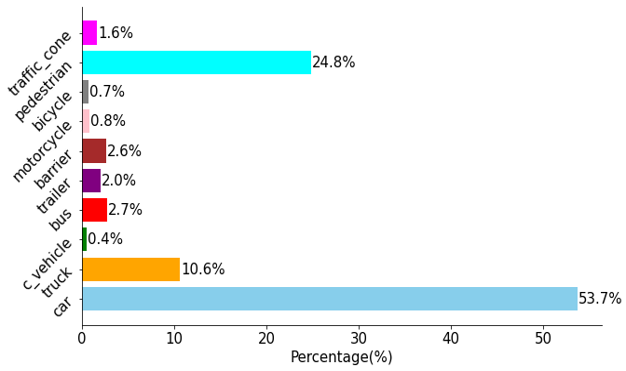}
            \label{fig:b}}
  \caption{Category distribution of referred objects in the Talk2Car-3D testing set: (a).\textit{"unique"} samples. (b).\textit{"multiple"} samples.}
  \label{category}
\end{figure}

In our experiments, we observe a strange phenomenon where the accuracy of simpler samples labeled as \textit{unique} is actually lower.
Fig.\ref{category} illustrates the category distribution of referred objects in the Talk2Car-3D testing set, revealing a significant bias between the two attribute labels. 
Cars are the most common objects in the dataset, yet they don’t appear in \textit{“unique”} samples. 
Additionally, we find that the average distance for \textit{“unique”} samples is 24.8 meters, higher than the 21.6 meters for \textit{“multiple”} samples. These factors pose greater challenges for the former, resulting in decreased model accuracy.

\clearpage


\newpage
\section*{NeurIPS Paper Checklist}


\begin{enumerate}

\item {\bf Claims}
    \item[] Question: Do the main claims made in the abstract and introduction accurately reflect the paper's contributions and scope?
    \item[] Answer: \answerYes{} 
    \item[] Justification: We clearly articulated the contributions and innovations of this paper in the abstract and introduction.

\item {\bf Limitations}
    \item[] Question: Does the paper discuss the limitations of the work performed by the authors?
    \item[] Answer: \answerYes{} 
    \item[] Justification: We discuss the limitations of this work in Sec.\ref{5.2}, Sec.\ref{5.3}, and Sec.\ref{6}.

\item {\bf Theory Assumptions and Proofs}
    \item[] Question: For each theoretical result, does the paper provide the full set of assumptions and a complete (and correct) proof?
    \item[] Answer: \answerNA{} 
    \item[] Justification: This paper focuses on the human-LiDAR interaction in autonomous driving without involving theoretical assumptions and proofs.

    \item {\bf Experimental Result Reproducibility}
    \item[] Question: Does the paper fully disclose all the information needed to reproduce the main experimental results of the paper to the extent that it affects the main claims and/or conclusions of the paper (regardless of whether the code and data are provided or not)?
    \item[] Answer: \answerYes{} 
    \item[] Justification: We provide detailed information about the dataset in Sec.\ref{3} and App.\ref{a1}, and describe our method in Sec.\ref{4}. All implementation details are illustrated in Sec.\ref{5.1} and App.\ref{a21}.

\item {\bf Open access to data and code}
    \item[] Question: Does the paper provide open access to the data and code, with sufficient instructions to faithfully reproduce the main experimental results, as described in supplemental material?
    \item[] Answer: \answerNo{} 
    \item[] Justification: We will open-source all data, code, and checkpoints upon paper acceptance.

\item {\bf Experimental Setting/Details}
    \item[] Question: Does the paper specify all the training and test details (e.g., data splits, hyperparameters, how they were chosen, type of optimizer, etc.) necessary to understand the results?
    \item[] Answer: \answerYes{} 
    \item[] Justification: We provide all the training and test details in Sec.\ref{5.1} and App.\ref{a21}.

\item {\bf Experiment Statistical Significance}
    \item[] Question: Does the paper report error bars suitably and correctly defined or other appropriate information about the statistical significance of the experiments?
    \item[] Answer: \answerYes{} 
    \item[] Justification: We provide all experiment statistics in Sec.\ref{5.2}. All experimental results are the statistical averages of five trials.
    \begin{itemize}
        \item The answer NA means that the paper does not include experiments.
        \item The authors should answer "Yes" if the results are accompanied by error bars, confidence intervals, or statistical significance tests, at least for the experiments that support the main claims of the paper.
        \item The factors of variability that the error bars are capturing should be clearly stated (for example, train/test split, initialization, random drawing of some parameter, or overall run with given experimental conditions).
        \item The method for calculating the error bars should be explained (closed form formula, call to a library function, bootstrap, etc.)
        \item The assumptions made should be given (e.g., Normally distributed errors).
        \item It should be clear whether the error bar is the standard deviation or the standard error of the mean.
        \item It is OK to report 1-sigma error bars, but one should state it. The authors should preferably report a 2-sigma error bar than state that they have a 96\% CI, if the hypothesis of Normality of errors is not verified.
        \item For asymmetric distributions, the authors should be careful not to show in tables or figures symmetric error bars that would yield results that are out of range (e.g. negative error rates).
        \item If error bars are reported in tables or plots, The authors should explain in the text how they were calculated and reference the corresponding figures or tables in the text.
    \end{itemize}

\item {\bf Experiments Compute Resources}
    \item[] Question: For each experiment, does the paper provide sufficient information on the computer resources (type of compute workers, memory, time of execution) needed to reproduce the experiments?
    \item[] Answer: \answerYes{} 
    \item[] Justification: We provide the information about computing resources in App.\ref{a21}.
    
\item {\bf Code Of Ethics}
    \item[] Question: Does the research conducted in the paper conform, in every respect, with the NeurIPS Code of Ethics \url{https://neurips.cc/public/EthicsGuidelines}?
    \item[] Answer: \answerYes{} 
    \item[] Justification: This research paper complies with the NeurIPS Code of Ethics.

\item {\bf Broader Impacts}
    \item[] Question: Does the paper discuss both potential positive societal impacts and negative societal impacts of the work performed?
    \item[] Answer: \answerYes{} 
    \item[] Justification: We discuss the potential social impacts in Sec.\ref{1} and \ref{6}.

\item {\bf Safeguards}
    \item[] Question: Does the paper describe safeguards that have been put in place for responsible release of data or models that have a high risk for misuse (e.g., pretrained language models, image generators, or scraped datasets)?
    \item[] Answer: \answerNA{} 
    \item[] Justification: This article does not address this point.

\item {\bf Licenses for existing assets}
    \item[] Question: Are the creators or original owners of assets (e.g., code, data, models), used in the paper, properly credited and are the license and terms of use explicitly mentioned and properly respected?
    \item[] Answer: \answerYes{} 
    \item[] Justification: We have secured proper attribution and licensing for all intellectual property used in this paper.

\item {\bf New Assets}
    \item[] Question: Are new assets introduced in the paper well documented and is the documentation provided alongside the assets?
    \item[] Answer: \answerYes{} 
    \item[] Justification: We provide comprehensive documentation and details about the new assets in the paper. 

\item {\bf Crowdsourcing and Research with Human Subjects}
    \item[] Question: For crowdsourcing experiments and research with human subjects, does the paper include the full text of instructions given to participants and screenshots, if applicable, as well as details about compensation (if any)? 
    \item[] Answer: \answerNA{} 
    \item[] Justification: This paper does not involve any crowdsourcing or research with human subjects.

\item {\bf Institutional Review Board (IRB) Approvals or Equivalent for Research with Human Subjects}
    \item[] Question: Does the paper describe potential risks incurred by study participants, whether such risks were disclosed to the subjects, and whether Institutional Review Board (IRB) approvals (or an equivalent approval/review based on the requirements of your country or institution) were obtained?
    \item[] Answer: \answerNA{} 
    \item[] Justification: This paper does not involve any crowdsourcing or research with human subjects.

\end{enumerate}

\end{document}